\pdfoutput=1
\documentclass[11pt,a4paper]{article}
\usepackage[hyperref]{emnlp2020_arxiv}
\usepackage{times}
\usepackage{latexsym}

\usepackage{microtype}
\aclfinalcopy

\usepackage{adjustbox}
\usepackage{graphicx}
\usepackage[normalem]{ulem}

\def\Snospace~{\S{}}

\usepackage{booktabs}
\usepackage{placeins}
\usepackage{amssymb}
\usepackage{amsmath}
\usepackage{pifont}
\newcommand{\cmark}{\ding{51}}
\newcommand{\xmark}{\ding{55}}
\usepackage{bbm}
\usepackage{soul}
\definecolor{darkgreen}{HTML}{006400}
\definecolor{darkgrey}{HTML}{AFAFAF}

\newcommand{\mcL}{\mathcal{L}}
\newcommand{\mcV}{\mathcal{V}}
\newcommand\given{\,|\,}

\title{Simulated Multiple Reference Training
\\ Improves Low-Resource Machine Translation}

 \author{Huda Khayrallah$~~~~$ Brian Thompson $~~~~$\bf Matt Post \and Philipp Koehn \\
 Johns Hopkins University\\
   {\tt \{huda, brian.thompson, phi\}@jhu.edu}, {\tt post@cs.jhu.edu}
 }
\date{}

\begin{document}
\setlength{\abovedisplayskip}{1pt}
\setlength{\belowdisplayskip}{5pt}
\setlength{\abovedisplayshortskip}{1pt}
\setlength{\belowdisplayshortskip}{5pt}
\maketitle
\begin{abstract}
Many valid translations exist for a given sentence, yet machine translation (MT) is trained with a single reference translation, exacerbating data sparsity in low-resource settings. We introduce Simulated Multiple Reference Training (SMRT), a novel MT training method that approximates the full space of possible translations by \emph{sampling} a paraphrase of the reference sentence from a paraphraser and training the MT model to predict the paraphraser's \emph{distribution} over possible tokens. We demonstrate the effectiveness of SMRT in low-resource settings when translating to English, with  improvements of 1.2 to 7.0 BLEU. We also find SMRT is complementary to back-translation.

\end{abstract}
\FloatBarrier
\section{Introduction}
Variability and expressiveness are  core features of language, and they extend to translation as well. \citet{dreyer-marcu-2012-hyter} showed that naturally occurring sentences have \emph{billions} of valid translations.
Despite this variety, machine translation (MT) models are optimized toward a single translation of each sentence in the training corpus. 
Training a high resource MT model on millions of sentence pairs likely exposes it to similar sentences translated different ways,
but training a low-resource MT model with a single translation for each sentence (out of potentially billions) exacerbates
data sparsity.

We hypothesize that the discrepancy between linguistic diversity and standard single-reference training hinders MT performance. 
This was previously impractical to address, since obtaining multiple human translations of training data is typically not feasible.
However, recent neural sentential paraphrasers produce fluent, meaning-preserving English paraphrases. 
We introduce a novel method that incorporates such a paraphraser directly in the training objective, and uses it to simulate the full space of translations.

We demonstrate the effectiveness of our method on two corpora from the low-resource MATERIAL program, and on  bitext from GlobalVoices.  We release data \& code:  \href{http://data.statmt.org/smrt}{\texttt{data.statmt.org/smrt}}

\begin{figure}
    \centering
    \includegraphics{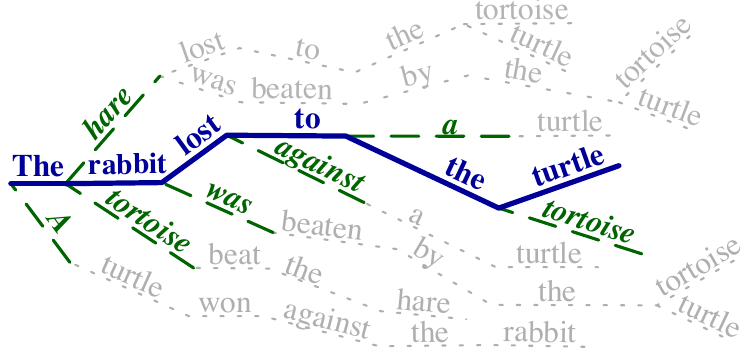}

    \caption{Some \textcolor{darkgrey}{\dotuline{possible paraphrases}} of the original reference, `The tortoise beat the hare,' for the Dutch source sentence, `De schildpad versloeg de haas.' A \textcolor{darkblue}{\uline{\textbf{sampled path}}} and some of the other   \textcolor{darkgreen}{\textbf{\textit{\dashuline{tokens also considered in the training objective}}}} are highlighted.}
\label{fig:paraphrase_lattice}
\end{figure}

 \section{Method}

We propose Simulated Multiple Reference Training (SMRT), 
which uses a paraphraser to approximate the full space of possible translations,
since explicitly training on billions of possible translations per sentence is intractable.

In standard neural MT training, the reference is: 
(1)~used in the training objective; and 
(2)~conditioned on as the previous target token.\footnote{In autoregressive NMT inference, predictions condition on the previous target tokens. In training, predictions typically condition on the previous tokens in the reference, not the model's output \cite[teacher forcing;][]{doi:10.1162/neco.1989.1.2.270}.}
We approximate the full space of possible translations by: 
(1)~training the MT model to predict the \emph{distribution} over possible tokens from the paraphraser at each time step;
 and 
(2)~\emph{sampling} the previous target token from the  paraphraser distribution. 
\autoref{fig:paraphrase_lattice} shows an example of possible paraphrases and highlights a sampled path and some of the other tokens used in the training objective distribution. 

We review the standard NLL training objective, and then introduce our proposed objective.
\paragraph{NLL Objective} The standard negative log likelihood (NLL) training objective in NMT, for the $i^{\textit{th}}$ target word in the  reference $y$ is:\textcolor{white}{$\mathbbm{1}$}
\begin{align}\label{eq:nll}
\mcL_{\text{NLL}}= - \sum_{v \in \mcV}^{} \Big[  \mathbbm{1}\{&y_i=v\} \\ \nonumber
\times \; \log\;  p_{\textsc{mt}}(&y_i=v \given x,y_{j < i}) \Big]
\end{align}
where $\mcV$ is the vocabulary, 
$\mathbbm{1}\{\cdot\}$ is the indicator function, 
and $p_{\textsc{mt}}$ is the MT output distribution (conditioned on the source $x$, and on the previous tokens in the reference $y_{j<i}$).
\autoref{eq:nll} computes the cross-entropy between the MT model's distribution and the one-hot reference. 
\paragraph{Proposed Objective}
We compute the cross entropy between the distribution of the MT model and the distribution from a paraphraser conditioned on the original reference:
\begin{align}\label{eq:para}
\mcL_{\text{\text{SMRT}}} = - \sum_{v \in \mcV}^{} \Big[ \;  p_{\text{para}}&(y'_i=v \given y,y'_{j<i}) \\ \nonumber
\times \; \log \;  p_{\textsc{mt}}&(y'_i=v \given x,y'_{j<i})\Big] 
\end{align}
where $y'$ is a paraphrase of the 
original reference $y$. 
$p_{\text{para}}$ is the output distribution from the paraphraser (conditioned on the reference $y$ and the previous tokens in the sentence produced by the paraphraser $y'_{j<i}$). 
$p_{\textsc{mt}}$ is the MT output distribution (conditioned on the source sentence, $x$ and the previous tokens in the sentence produced by the paraphraser, $y'_{j<i}$).
At each time step we sample a target token $y'_{i}$ from the paraphraser's output distribution 
to cover the space of translations.
We condition on the sampled $y'_{i-1}$ as the previous target token for both the MT model and paraphraser.

For a visualization see \autoref{fig:paraphrase_lattice}, which shows \textcolor{darkgrey}{\dotuline{possible paraphrases}} of the reference, `The tortoise beat the hare.'   
The paraphraser and MT model condition on the \textcolor{darkblue}{\uline{\textbf{paraphrase ($y'$)}}} as the previous output. The \textcolor{darkblue}{\uline{\textbf{paraphrase ($y'$)}}} and the rest of the \textcolor{darkgreen}{\textbf{\textit{\dashuline{tokens  in the paraphraser's distribution}}}} make up $p_{\textsc{para}}$, which is used to compute $\mcL_{\text{SMRT}}$.

\begin{table*}[ht]
\centering
\begin{tabular}{l|rrrrrrrrrr|rc}
 \toprule
dataset & \multicolumn{10}{c}{GlobalVoices} & \multicolumn{2}{|c}{MATERIAL} 
\\
\midrule
 * $\rightarrow$ en & \multicolumn{1}{c}{hu} & \multicolumn{1}{c}{id} & \multicolumn{1}{c}{cs}  & \multicolumn{1}{c}{sr}  & \multicolumn{1}{c}{ca}  & \multicolumn{1}{c}{sw}  & \multicolumn{1}{c}{nl}  & \multicolumn{1}{c}{pl}  & \multicolumn{1}{c}{mk}  & \multicolumn{1}{c}{ar} &\multicolumn{1}{|c}{sw} & \multicolumn{1}{c}{tl} 
         \\
train lines  & \multicolumn{1}{c}{8k} & \multicolumn{1}{c}{8k} & \multicolumn{1}{c}{11k} & \multicolumn{1}{c}{14k} & \multicolumn{1}{c}{15k} & \multicolumn{1}{c}{24k} & \multicolumn{1}{c}{32k} & \multicolumn{1}{c}{40k} & \multicolumn{1}{c}{44k} & \multicolumn{1}{c}{47k}   &   \multicolumn{1}{|c}{19k} & \multicolumn{1}{c}{46k}  
                                                \\ \midrule
baseline  & 2.3          & 5.3           & 3.4          & 11.8          & 16.0          & 17.9          & 22.2          & 16.0          & 27.0          & 12.7          & 37.8          & 32.5          \\
this work & \textbf{5.4} & \textbf{12.3} & \textbf{6.6} & \textbf{16.1} & \textbf{20.0} & \textbf{20.5} & \textbf{24.8} & \textbf{18.0} & \textbf{28.2} & \textbf{14.9} & \textbf{39.0} & \textbf{33.7} \\
\midrule
$\Delta$  & +3.1          & +7.0           & +3.2          & +4.3           & +4.0           & +2.6           & +2.6           & +2.0           & +1.2           & +2.2           & +1.2           & +1.2          
           \\
\bottomrule
\end{tabular}
\caption{BLEU scores on the test set. We \textbf{bold} the best value; all improvements are statistically significant at the 95\% confidence level. `train lines' indicates amount of training bitext.
}
\label{tab:mainresults}
\end{table*}
\section{Experimental Setup}
\label{sec:experiment}
\subsection{Paraphraser}
\label{sec:pp}
For use as an English paraphraser,
we train a Transformer model \cite{transformer} in \textsc{fairseq} \cite{ott-etal-2019-fairseq} with an
 8-layer encoder and decoder,  $1024$ dimensional embeddings, $16$ encoder and decoder attention heads, and $0.3$ dropout. 
We optimize using Adam \cite{kingma2014adam}. 
We train on \textsc{ParaBank2} \cite{hu-etal-2019-large}, an English paraphrase dataset.\footnote{\citeauthor{hu-etal-2019-large} released a trained \textsc{Sockeye} paraphraser but we implement our method in \textsc{fairseq}.}
\textsc{ParaBank2} was generated by training an MT system on CzEng 1.7 (a Czech$-$English bitext with over $50$ million lines \cite{CzEng}), re-translating the Czech training sentences, and pairing the English output with the original English translation.  

\subsection{NMT models}
\label{sec:nmt_models}
We train Transformer NMT models in \textsc{fairseq} using the \textsc{flores} low-resource benchmark parameters \cite{guzman-etal-2019-flores}: $5$-layer encoder and decoder, $512$-dimensional embeddings, and $2$ encoder and decoder attention heads. We regularize with $0.2$ label smoothing and $0.4$ dropout. 
We optimize using Adam with a learning rate of $10^{-3}$. 
We train for 200 epochs, and select the best checkpoint based on validation set perplexity. 
We translate with a beam size of $5$.
For our method we use the proposed objective $\mcL_{\text{SMRT}}$ with probability $p=0.5$ and standard $\mcL_{\text{NLL}}$ on the original reference with probability $1-p$.
We sample from only the 100 highest probability vocabulary items at a given time step when sampling from the paraphraser distribution to avoid very unlikely tokens \cite{fan-etal-2018-hierarchical}. 

Using our English paraphraser, we aim to demonstrate improvements in low-resource settings, since these remain a challenge in NMT \cite{koehn-knowles-2017-six,sennrich-zhang-2019-revisiting}. 
We use Tagalog (tl) to English and Swahili (sw) to English bitext from the MATERIAL low-resource program \cite{rubino-2018-keynote}.
We also report results on MT bitext from GlobalVoices,
a non-profit news site that publishes in $53$ languages.\footnote{We use v2017q3 released on Opus \cite[][\href{http://opus.nlpl.eu/GlobalVoices.php}{\texttt{opus.nlpl.eu/GlobalVoices.php}}]{tiedemann-2012-parallel}.}
We evaluate on the 10 lowest-resource settings that have at least 10,000 lines of parallel text with English: Hungarian (hu), Indonesian (id), Czech (cs), Serbian (sr), Catalan (ca), Swahili (sw),\footnote{Swahili is in both. MATERIAL data is not widely available, so we separate them to keep GlobalVoices reproducible.} Dutch (nl), Polish (pl), Macedonian (mk), Arabic (ar).

We use 2,000 lines each for a validation set for model selection from checkpoints and a test set for reporting results. 
The approximate number of lines of training data is in the top of \autoref{tab:mainresults}.
We train an English SentencePiece model \cite{kudo-richardson-2018-sentencepiece}  on the paraphraser data, and apply it to the target (English) side of the MT bitext,
 so that the paraphraser and MT models have the same output vocabulary. We also train SentencePiece models on the source-side of the bitexts.
We use a subword vocabulary size of 4,000 for each. 

\section{Results }
\label{sec:results}
Results are shown in \autoref{tab:mainresults}. Our method improves over the baseline in all settings, by between 1.2 and 7.0 BLEU (all statistically significant at the  95\% confidence level \cite{koehn-2004-statistical}).\footnote{All BLEU scores are SacreBLEU \cite{post-2018-call}.} 
We see larger improvements for lower-resource corpora. 

\section{Analysis}
\label{sec:analysis}
We analyze
our method to explore:
(1)~how it performs at a various resource levels; (2)~how it combines with back-translation; (3)~how the different components of the method impact performance; and (4)~how it compares to sequence-level paraphrastic data augmentation. 

\begin{table*}[ht]
 \centering
\addtolength{\tabcolsep}{-1.5pt}
\scalebox{.955}{%
\begin{tabular}{l|rrrrrrrrrr|rc}
 \toprule
dataset & \multicolumn{10}{c}{GlobalVoices} & \multicolumn{2}{|c}{ MATERIAL} 
\\
\midrule
 * $\rightarrow$ en & \multicolumn{1}{c}{hu} & \multicolumn{1}{c}{id} & \multicolumn{1}{c}{cs}  & \multicolumn{1}{c}{sr}  & \multicolumn{1}{c}{ca}  & \multicolumn{1}{c}{sw}  & \multicolumn{1}{c}{nl}  & \multicolumn{1}{c}{pl}  & \multicolumn{1}{c}{mk}  & \multicolumn{1}{c}{ar} &\multicolumn{1}{|c}{sw} & \multicolumn{1}{c}{tl} 
         \\
train lines  & \multicolumn{1}{c}{8k} & \multicolumn{1}{c}{8k} & \multicolumn{1}{c}{11k} & \multicolumn{1}{c}{14k} & \multicolumn{1}{c}{15k} & \multicolumn{1}{c}{24k} & \multicolumn{1}{c}{32k} & \multicolumn{1}{c}{40k} & \multicolumn{1}{c}{44k} & \multicolumn{1}{c}{47k}   &   \multicolumn{1}{|c}{19k} & \multicolumn{1}{c}{46k}  
\\ \midrule
baseline                      & 2.3          & 5.3           & 3.4          & 11.8          & 16.0          & 17.9          & 22.2          & 16.0          & 27.0          & 12.7          & 37.8          & 32.5          \\
baseline w/ back-translation  & 2.8          & 7.1           & 4.6          & 17.6          & 20.1          & 20.7          & 26.9          & 19.3          & 29.1          & 16.0          & 38.8          & 33.0          \\
\midrule
this work                     & \textbf{5.4} & 12.3          & \textbf{6.6} & 16.1          & 20.0          & 20.5          & 24.8          & 18.0          & 28.2          & 14.9          & 39.0          & \textbf{33.7} \\
this work w/ back-translation & 4.9          & \textbf{12.8} & \textbf{6.6} & \textbf{19.6} & \textbf{23.4} & \textbf{23.0} & \textbf{27.5} & \textbf{20.2} & \textbf{29.7} & \textbf{16.8} & \textbf{39.3} & \textbf{33.7}\\
\bottomrule
\end{tabular}
}
\caption{
Comparison between back-translation and this work.
We \textbf{bold} the best BLEU score on the test set, as well as any result where the difference from it is not statistically significant at the 95\% confidence level.
}
\label{tab:bt}
\end{table*}

\subsection{MT Data Ablation}
In order to better understand how our method performs across data sizes on the same corpus, we ablate Bengali-English bitext from GlobalVoices.

\begin{figure}
    \centering
    \includegraphics{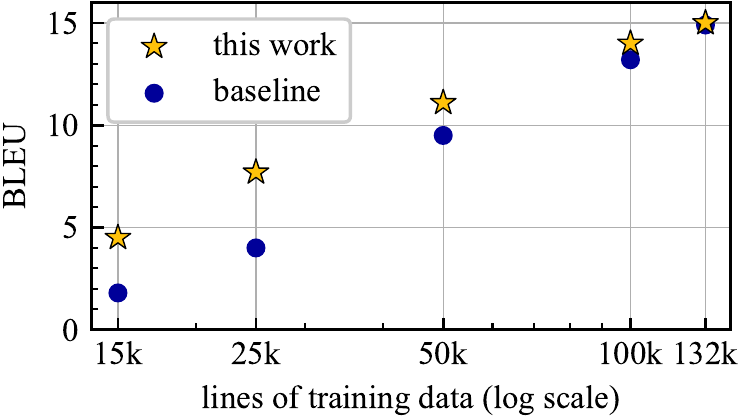}
    \caption{Bengali-English data ablation. Improvements of 2.7, 3.7, 1.6, and 0.8 BLEU at the 15k, 25k, 50k, and 100k subsets are statistically significant.}
    \label{fig:subsets}
\end{figure}

\autoref{fig:subsets} plots the performance of our method and the baseline against the log of the data amount. 
Our improvements of 2.7, 3.7, 1.6, and 0.8 BLEU at the 15k, 25k, 50k, and 100k subsets are statistically significant at the 95\% confidence level; the 0.1 improvement for the full 132k data amount is not. Similar to \autoref{tab:mainresults}, we see larger improvements in lower-resource ablations.

\subsection{Back-translation}
\label{sec:bt}
Back-translation \cite{sennrich-etal-2016-improving} is the de facto method for incorporating non-parallel data in NMT, so we investigate how our method interacts with it.
\autoref{tab:bt} shows the results for  back-translation, our work, and the combination of both.\footnote{We use a 1:1 ratio of bitext to back-translated bitext.
We use newscrawl2016 (\href{http://data.statmt.org/news-crawl/}{\texttt{data.statmt.org/news-crawl}}) as monolingual text. 
When combining with our work, we run our method on both the original and back-translation data.} 
Adding our method to back-translation improves results by an additional 0.5 to 5.7 BLEU.\footnote{\label{statsig}All statistically significant at the 95\% confidence level.}

For all language pairs, the best performance is achieved by our method combined with back-translation, or our method alone. 
For 9 of 12 corpora, back-translation and our proposed method are complementary, with improvements of 1.2 to 7.8 BLEU\textsuperscript{\ref{statsig}} over the baseline when combining the two. 
For cs-en and tl-en, adding back-translation to our method does not change BLEU.
In the lowest-resource setting (hu-en) our method alone outperforms the baseline by 3.1 BLEU, but adding back-translation reduces the improvement by 0.5 BLEU.  

\begin{table*}[ht]
\centering
\addtolength{\tabcolsep}{-2.7pt}
\begin{tabular}{ccl|rrrrrrrrrr|rc}
 \toprule
& &\multicolumn{1}{l|}{\phantom{(3)} dataset} & \multicolumn{10}{c}{GlobalVoices} & \multicolumn{2}{|c}{MATERIAL} 
\\
\midrule
 dist. & paraphrase & \multicolumn{1}{l|}{\phantom{(3)} * $\rightarrow$ en} & \multicolumn{1}{c}{hu} & \multicolumn{1}{c}{id} & \multicolumn{1}{c}{cs}  & \multicolumn{1}{c}{sr}  & \multicolumn{1}{c}{ca}  & \multicolumn{1}{c}{sw}  & \multicolumn{1}{c}{nl}  & \multicolumn{1}{c}{pl}  & \multicolumn{1}{c}{mk}  & \multicolumn{1}{c}{ar} &\multicolumn{1}{|c}{sw} & \multicolumn{1}{c}{tl} 
         \\
  loss& sampling &\multicolumn{1}{l|}{\phantom{(3)} train lines} & \multicolumn{1}{c}{8k} & \multicolumn{1}{c}{8k} & \multicolumn{1}{c}{11k} & \multicolumn{1}{c}{14k} & \multicolumn{1}{c}{15k} & \multicolumn{1}{c}{24k} & \multicolumn{1}{c}{32k} & \multicolumn{1}{c}{40k} & \multicolumn{1}{c}{44k} & \multicolumn{1}{c}{47k}   &   \multicolumn{1}{|c}{19k} & \multicolumn{1}{c}{46k} \\ 
\midrule
  \xmark &  n/a  &\phantom{(3)} baseline     & 2.3          & 5.3           & 3.4          & 11.8          & 16.0          & 17.9          & 22.2          & 16.0          & 27.0          & 12.7          & 37.8          & 32.5          \\\midrule
 \xmark &  \xmark &  (1) & 2.9          & 8.8           & 4.6          & 14.5          & 17.8          & 19.2          & 23.4          & 17.6          & 27.0            & 14.2          & 35.7          & 29.9          \\
 \xmark &  \cmark   &   (2)  & 5.1          & 11.6          & \textbf{6.5} & 15.6          & \textbf{19.7} & \textbf{20.2} & 24.4          & \textbf{18.1} & \textbf{27.9} & \textbf{15.0}   & 38.1          & 32.0         \\
 \cmark &  \xmark  &  (3) & 4.0            & 10.5          & \textbf{6.5} & 15.2          & 18.8          & 19.8          & 23.9          & \textbf{18.0}   & \textbf{27.6} & 14.4          & 37.6          & 31.6          \\
\midrule
  \cmark &  \cmark     & (4) this work     & \textbf{5.4} & \textbf{12.3} & \textbf{6.6} & \textbf{16.1} & \textbf{20.0} & \textbf{20.5} & \textbf{24.8} & \textbf{18.0} & \textbf{28.2} & \textbf{14.9} & \textbf{39.0} & \textbf{33.7}\\
\bottomrule
\end{tabular}
\caption{ We compare four conditions to the baseline: 
(1) paraphrasing the reference, without sampling or the distribution in the loss; (2) sampling from the paraphraser in the training objective, without the distribution; (3) using the distribution  in the training objective, without sampling; and (4) the proposed method.
 We \textbf{bold} the best test set BLEU score, and others where the difference is not statistically significant at the 95\% confidence level. 
}
\label{tab:ablate}
\end{table*}

\begin{table*}[ht]
\centering
\addtolength{\tabcolsep}{-1.5pt}
\begin{tabular}{l|rrrrrrrrrr|rc}
 \toprule
dataset & \multicolumn{10}{c}{GlobalVoices} & \multicolumn{2}{|c}{MATERIAL} 
\\
\midrule
 * $\rightarrow$ en & \multicolumn{1}{c}{hu} & \multicolumn{1}{c}{id} & \multicolumn{1}{c}{cs}  & \multicolumn{1}{c}{sr}  & \multicolumn{1}{c}{ca}  & \multicolumn{1}{c}{sw}  & \multicolumn{1}{c}{nl}  & \multicolumn{1}{c}{pl}  & \multicolumn{1}{c}{mk}  & \multicolumn{1}{c}{ar} &\multicolumn{1}{|c}{sw} & \multicolumn{1}{c}{tl} 
         \\
train lines  & \multicolumn{1}{c}{8k} & \multicolumn{1}{c}{8k} & \multicolumn{1}{c}{11k} & \multicolumn{1}{c}{14k} & \multicolumn{1}{c}{15k} & \multicolumn{1}{c}{24k} & \multicolumn{1}{c}{32k} & \multicolumn{1}{c}{40k} & \multicolumn{1}{c}{44k} & \multicolumn{1}{c}{47k}   &   \multicolumn{1}{|c}{19k} & \multicolumn{1}{c}{46k}  
                                                \\ \midrule
baseline  & 2.3          & 5.3           & 3.4          & 11.8          & 16.0          & 17.9          & 22.2          & 16.0                   & 27.0                   & 12.7                   & 37.8          & 32.5     \\
\midrule
beam-search paraphrase      & 2.6 & 8.7  & 4.7 & 13.5 & 16.3 & 18.4 & 22.6          & 16.6       & 26.6                   & 12.2                   & 35.9          & 29.4     \\
greedy paraphrase  &3.2 & 9.4  & 4.6 & 14.8 & 18.3 & 19.6 & 24.4 & \textbf{18.0} & \textbf{27.5} & \textbf{14.7} & 35.8          & 30.3   \\
 sampled paraphrase    & 2.8 & 8.0  & 5.1 & 13.9 & 16.8 & 19.5 & 23.9 & 17.6        & \textbf{27.6} & 14.2      & 37.2          & 31.6     \\
\midrule
this work & \textbf{5.4} & \textbf{12.3} & \textbf{6.6} & \textbf{16.1} & \textbf{20.0} & \textbf{20.5} & \textbf{24.8} & \textbf{18.0}          & \textbf{28.2}          & \textbf{14.9}          & \textbf{39.0} & \textbf{33.7} \\
\bottomrule
\end{tabular}
\caption{  We compare three ways of generating paraphrases for preprocessed data augmentation: beam search, greedy search, and sampling.
We \textbf{bold} the best BLEU score on the test set, as well as any result where the difference from it is not statistically significant at the 95\% confidence level. 
}
\label{tab:offline_}
\end{table*}

\subsection{Method Ablation}
 In \autoref{tab:ablate} we analyze the contributions of each part of our proposed method.
We compare four conditions to the baseline:\footnote{\label{ablationsetting}All use settings from \autoref{sec:nmt_models}: we use the original reference with $\mcL_{\text{NLL}}$ with $1-p=0.5$ probability, and when sampling we sample from the top $w=100$ tokens.} 
(1) paraphrasing the reference, without sampling or the distribution in the loss;\footnote{This is equivalent to $\mcL_{\text{NLL}}$ using a  paraphrase generated with greedy-search as the reference, see \autoref{sec:offline}.} (2) sampling from the paraphraser, without the distribution in the loss; (3) using the distribution in the training objective, without sampling the paraphrase; and (4) the proposed method.

We find that sampling is particularly important to the success for the method; removing it significantly degrades  performance in all but 3 language pairs. Since we sample a paraphrase each batch, this exposes the model to a wide variety of different paraphrases. Using the distribution in the loss function is also beneficial, particularly for the lower resource settings and in the MATERIAL corpora. 

\subsection{Sequence-Level Paraphrastic Data Augmentation}
\label{sec:offline}
As a contrastive experiment, we use the paraphraser to generate additional target-side data for use in data augmentation. 
For each target sentence ($y$) in the training data, we generate a paraphrase ($y'$). 
We then concatenate the original  source-target pairs $(x,y)$ with the paraphrased pairs $(x,y')$ and perform standard standard $\mcL_{\text{NLL}}$ training. 
We consider 3 methods for generating paraphrases: beam search (beam of 5), greedy search, sampling (top-100 sampling). Greedy search tends to work best: see \autoref{tab:offline_}. It improves over the baseline for the 10 Global Voices datasets, but not for the two MATERIAL ones.
Overall, our proposed method is more effective than this contrastive method.
We hypothesize this is due to the wider variety of paraphrases SMRT introduces by sampling and training toward the full distribution from the paraphraser. 

\section{Related Work}
\label{sec:related_work}
\paragraph{Knowledge Distillation}
Our proposed objective is similarly structured to word-level knowledge distillation \cite[KD;][]{hinton2015distilling,kim-rush-2016-sequence},
where a student model is trained to match the output distribution of a teacher model. 
Paraphrasing as preprocessed data augmentation, as discussed in \autoref{sec:offline}, is similarly analogous to sequence-level knowledge distillation \cite{kim-rush-2016-sequence}.

In typical KD both the student and teacher models are translation models trained on the same data, have the same input and output languages, and use the original reference for the previous token. 
In contrast, our teacher model is a paraphraser, which takes as input the original reference sentence (in the target language), with the sampled paraphrase as the previous token.
KD is usually used to train smaller models and does not typically incorporate additional data sources, though it has been used for domain adaptation \cite{dakwale2017fine, khayrallah-etal-2018-regularized}. 
\paragraph{Paraphrasing in MT}
\citet{hu-etal-2019-improved} present case studies on paraphrastic data augmentation for NLP tasks, including NMT. 
They use sequence-level augmentation with heuristic constraints on the model's output. 
SMRT differs in that we train toward the paraphraser \emph{distribution}, and we \emph{sample} from the distribution rather than using heuristics.

\citet{wieting-etal-2019-beyond} used a paraphrastic-similarity metric for minimum risk training \cite[MRT;][]{shen-etal-2016-minimum} in NMT.
They note MRT is slow, and, following prior work, use it for fine-tuning after NLL training. While our method is about $3$ times slower than standard $\mcL_{\text{NLL}}$, this is not prohibitive in low-resource conditions. 

 \label{sec:pp_for_da}
Paraphrasing was also used for statistical MT, including:  \emph{source-side} phrase table augmentation \cite{Callison-Burch:2006:ISM:1220835.1220838,Marton:2009:ISM:1699510.1699560}, and generation of additional references for tuning \cite{madnani-etal-2007-using,madnaniamta08}.

\paragraph{Data Augmentation in NMT}
Back-translation (BT) translates target-language monolingual text to create synthetic source sentences \cite{sennrich-etal-2016-improving}.
BT needs a reverse translation model for each \emph{language pair}.
In contrast, we need a paraphraser for each \emph{target language}.
\citet{zhou-etal-2019-handling} found BT is harmful in some low-resource settings, but a strong paraphraser can be trained as long as the target language is sufficiently high resource. 
 
\citet{fadaee-etal-2017-data} insert rare words in novel contexts in the existing bitext, using automatic word alignment and a language model. 
 RAML \cite{NIPS2016_6547} and SwitchOut \cite{wang-etal-2018-switchout} 
 randomly replace words others from the vocabulary. 
In contrast to random or targeted word replacement, we generate semantically similar sentential paraphrases.
\paragraph{Label Smoothing}
Label smoothing (which we use when training with $\mcL_{\text{NLL}}$) spreads probability mass over all non-reference tokens equally \cite{labelsmooth}; $\mcL_{\text{SMRT}}$ 
places higher probability on semantically plausible tokens. 

\section{Conclusion}
We present Simulated Multiple Reference Training (SMRT), which significantly improves performance in low-resource settings---by 1.2 to 7.0 BLEU---and is complementary to back-translation. 

Neural paraphrasers are rapidly improving \cite{wieting-etal-2017-learning,wieting-etal-2019-simple,li-etal-2018-paraphrase,wieting-gimpel-2018-paranmt,hu-etal-2019-improved,parabank,hu-etal-2019-large}, and the concurrently released \textsc{Prism} multi-lingual paraphraser \cite{thompson-post-2020-automatic,thompson-post-2020-paraphrase} has coverage of 39 languages
 and outperforms prior work in English paraphrasing. As paraphrasing continues to improve and cover more languages, we are optimistic SMRT will provide larger improvements across the board---including for higher-resource MT and for additional target languages beyond English.
\section*{Acknowledgments}
Brian Thompson is supported 
by the National Defense Science and Engineering Graduate (NDSEG) Fellowship.
This material is based upon work supported by the United States Air Force under Contract No. FA8750‐19‐C‐0098, Learning with Less Labels.  Any opinions, findings, and conclusions or recommendations expressed in this material are those of the authors and do not necessarily reflect the views of the United States Air Force/DARPA.

\bibliography{anthology,extra} 

\begin{thebibliography}{40}
\expandafter\ifx\csname natexlab\endcsname\relax\def\natexlab#1{#1}\fi

\bibitem[{Bojar et~al.(2016)Bojar, Du{\v{s}}ek, Kocmi, Libovick{\'y},
  Nov{\'a}k, Popel, Sudarikov, and Vari{\v{s}}}]{CzEng}
Ond{\v{r}}ej Bojar, Ond{\v{r}}ej Du{\v{s}}ek, Tom Kocmi, Jind{\v{r}}ich
  Libovick{\'y}, Michal Nov{\'a}k, Martin Popel, Roman Sudarikov, and
  Du{\v{s}}an Vari{\v{s}}. 2016.
\newblock Czeng 1.6: Enlarged {Czech-English} parallel corpus with processing
  tools dockered.
\newblock In \emph{Text, Speech, and Dialogue}, pages 231--238, Cham. Springer
  International Publishing.

\bibitem[{Callison-Burch et~al.(2006)Callison-Burch, Koehn, and
  Osborne}]{Callison-Burch:2006:ISM:1220835.1220838}
Chris Callison-Burch, Philipp Koehn, and Miles Osborne. 2006.
\newblock \href {https://doi.org/10.3115/1220835.1220838} {Improved statistical
  machine translation using paraphrases}.
\newblock In \emph{Proceedings of the Main Conference on Human Language
  Technology Conference of the North American Chapter of the Association of
  Computational Linguistics}, HLT-NAACL '06, pages 17--24, Stroudsburg, PA,
  USA. Association for Computational Linguistics.

\bibitem[{Dakwale and Monz(2017)}]{dakwale2017fine}
Praveen Dakwale and Christof Monz. 2017.
\newblock Fine-tuning for neural machine translation with limited degradation
  across in-and out-of-domain data.
\newblock \emph{Proceedings of the XVI Machine Translation Summit}, page 117.

\bibitem[{Dreyer and Marcu(2012)}]{dreyer-marcu-2012-hyter}
Markus Dreyer and Daniel Marcu. 2012.
\newblock \href {https://www.aclweb.org/anthology/N12-1017} {{H}y{TER}:
  Meaning-equivalent semantics for translation evaluation}.
\newblock In \emph{Proceedings of the 2012 Conference of the North {A}merican
  Chapter of the Association for Computational Linguistics: Human Language
  Technologies}, pages 162--171, Montr{\'e}al, Canada. Association for
  Computational Linguistics.

\bibitem[{Fadaee et~al.(2017)Fadaee, Bisazza, and Monz}]{fadaee-etal-2017-data}
Marzieh Fadaee, Arianna Bisazza, and Christof Monz. 2017.
\newblock \href {https://doi.org/10.18653/v1/P17-2090} {Data augmentation for
  low-resource neural machine translation}.
\newblock In \emph{Proceedings of the 55th Annual Meeting of the Association
  for Computational Linguistics (Volume 2: Short Papers)}, pages 567--573,
  Vancouver, Canada. Association for Computational Linguistics.

\bibitem[{Fan et~al.(2018)Fan, Lewis, and Dauphin}]{fan-etal-2018-hierarchical}
Angela Fan, Mike Lewis, and Yann Dauphin. 2018.
\newblock \href {https://doi.org/10.18653/v1/P18-1082} {Hierarchical neural
  story generation}.
\newblock In \emph{Proceedings of the 56th Annual Meeting of the Association
  for Computational Linguistics (Volume 1: Long Papers)}, pages 889--898,
  Melbourne, Australia. Association for Computational Linguistics.

\bibitem[{Guzm{\'a}n et~al.(2019)Guzm{\'a}n, Chen, Ott, Pino, Lample, Koehn,
  Chaudhary, and Ranzato}]{guzman-etal-2019-flores}
Francisco Guzm{\'a}n, Peng-Jen Chen, Myle Ott, Juan Pino, Guillaume Lample,
  Philipp Koehn, Vishrav Chaudhary, and Marc{'}Aurelio Ranzato. 2019.
\newblock \href {https://doi.org/10.18653/v1/D19-1632} {The {FLORES} evaluation
  datasets for low-resource machine translation: {N}epali{--}{E}nglish and
  {S}inhala{--}{E}nglish}.
\newblock In \emph{Proceedings of the 2019 Conference on Empirical Methods in
  Natural Language Processing and the 9th International Joint Conference on
  Natural Language Processing (EMNLP-IJCNLP)}, pages 6100--6113, Hong Kong,
  China. Association for Computational Linguistics.

\bibitem[{Hinton et~al.(2015)Hinton, Vinyals, and Dean}]{hinton2015distilling}
Geoffrey Hinton, Oriol Vinyals, and Jeff Dean. 2015.
\newblock \href {http://arxiv.org/abs/1503.02531} {Distilling the knowledge in
  a neural network}.

\bibitem[{Hu et~al.(2019{\natexlab{a}})Hu, Khayrallah, Culkin, Xia, Chen, Post,
  and Van~Durme}]{hu-etal-2019-improved}
J.~Edward Hu, Huda Khayrallah, Ryan Culkin, Patrick Xia, Tongfei Chen, Matt
  Post, and Benjamin Van~Durme. 2019{\natexlab{a}}.
\newblock \href {https://doi.org/10.18653/v1/N19-1090} {Improved lexically
  constrained decoding for translation and monolingual rewriting}.
\newblock In \emph{Proceedings of the 2019 Conference of the North {A}merican
  Chapter of the Association for Computational Linguistics: Human Language
  Technologies, Volume 1 (Long and Short Papers)}, pages 839--850, Minneapolis,
  Minnesota. Association for Computational Linguistics.

\bibitem[{Hu et~al.(2019{\natexlab{b}})Hu, Rudinger, Post, and {Van
  Durme}}]{parabank}
J.~Edward Hu, Rachel Rudinger, Matt Post, and Benjamin {Van Durme}.
  2019{\natexlab{b}}.
\newblock \href {https://doi.org/10.1609/aaai.v33i01.33016521} {Para{B}ank:
  Monolingual bitext generation and sentential paraphrasing via
  lexically-constrained neural machine translation}.
\newblock In \emph{Proceedings of AAAI}.

\bibitem[{Hu et~al.(2019{\natexlab{c}})Hu, Singh, Holzenberger, Post, and
  Van~Durme}]{hu-etal-2019-large}
J.~Edward Hu, Abhinav Singh, Nils Holzenberger, Matt Post, and Benjamin
  Van~Durme. 2019{\natexlab{c}}.
\newblock \href {https://doi.org/10.18653/v1/K19-1005} {Large-scale, diverse,
  paraphrastic bitexts via sampling and clustering}.
\newblock In \emph{Proceedings of the 23rd Conference on Computational Natural
  Language Learning (CoNLL)}, pages 44--54, Hong Kong, China. Association for
  Computational Linguistics.

\bibitem[{Khayrallah et~al.(2018)Khayrallah, Thompson, Duh, and
  Koehn}]{khayrallah-etal-2018-regularized}
Huda Khayrallah, Brian Thompson, Kevin Duh, and Philipp Koehn. 2018.
\newblock \href {https://doi.org/10.18653/v1/W18-2705} {Regularized training
  objective for continued training for domain adaptation in neural machine
  translation}.
\newblock In \emph{Proceedings of the 2nd Workshop on Neural Machine
  Translation and Generation}, pages 36--44, Melbourne, Australia. Association
  for Computational Linguistics.

\bibitem[{Kim and Rush(2016)}]{kim-rush-2016-sequence}
Yoon Kim and Alexander~M. Rush. 2016.
\newblock \href {https://doi.org/10.18653/v1/D16-1139} {Sequence-level
  knowledge distillation}.
\newblock In \emph{Proceedings of the 2016 Conference on Empirical Methods in
  Natural Language Processing}, pages 1317--1327, Austin, Texas. Association
  for Computational Linguistics.

\bibitem[{Kingma and Ba(2015)}]{kingma2014adam}
Diederik~P. Kingma and Jimmy Ba. 2015.
\newblock \href {http://arxiv.org/abs/1412.6980} {Adam: {A} method for
  stochastic optimization}.
\newblock In \emph{3rd International Conference on Learning Representations,
  {ICLR} 2015, San Diego, CA, USA, May 7-9, 2015, Conference Track
  Proceedings}.

\bibitem[{Koehn(2004)}]{koehn-2004-statistical}
Philipp Koehn. 2004.
\newblock \href {https://www.aclweb.org/anthology/W04-3250} {Statistical
  significance tests for machine translation evaluation}.
\newblock In \emph{Proceedings of the 2004 Conference on Empirical Methods in
  Natural Language Processing}, pages 388--395, Barcelona, Spain. Association
  for Computational Linguistics.

\bibitem[{Koehn and Knowles(2017)}]{koehn-knowles-2017-six}
Philipp Koehn and Rebecca Knowles. 2017.
\newblock \href {https://doi.org/10.18653/v1/W17-3204} {Six challenges for
  neural machine translation}.
\newblock In \emph{Proceedings of the First Workshop on Neural Machine
  Translation}, pages 28--39, Vancouver. Association for Computational
  Linguistics.

\bibitem[{Kudo and Richardson(2018)}]{kudo-richardson-2018-sentencepiece}
Taku Kudo and John Richardson. 2018.
\newblock \href {https://doi.org/10.18653/v1/D18-2012} {{S}entence{P}iece: A
  simple and language independent subword tokenizer and detokenizer for neural
  text processing}.
\newblock In \emph{Proceedings of the 2018 Conference on Empirical Methods in
  Natural Language Processing: System Demonstrations}, pages 66--71, Brussels,
  Belgium. Association for Computational Linguistics.

\bibitem[{Li et~al.(2018)Li, Jiang, Shang, and Li}]{li-etal-2018-paraphrase}
Zichao Li, Xin Jiang, Lifeng Shang, and Hang Li. 2018.
\newblock \href {https://doi.org/10.18653/v1/D18-1421} {Paraphrase generation
  with deep reinforcement learning}.
\newblock In \emph{Proceedings of the 2018 Conference on Empirical Methods in
  Natural Language Processing}, pages 3865--3878, Brussels, Belgium.
  Association for Computational Linguistics.

\bibitem[{Madnani et~al.(2007)Madnani, Fazil~Ayan, Resnik, and
  Dorr}]{madnani-etal-2007-using}
Nitin Madnani, Necip Fazil~Ayan, Philip Resnik, and Bonnie Dorr. 2007.
\newblock \href {https://www.aclweb.org/anthology/W07-0716} {Using paraphrases
  for parameter tuning in statistical machine translation}.
\newblock In \emph{Proceedings of the Second Workshop on Statistical Machine
  Translation}, pages 120--127, Prague, Czech Republic. Association for
  Computational Linguistics.

\bibitem[{Madnani et~al.(2008)Madnani, Resnik, Dorr, and
  Schwartz}]{madnaniamta08}
Nitin Madnani, Philip Resnik, Bonnie~J. Dorr, and Richard Schwartz. 2008.
\newblock Are multiple reference translations necessary? investigating the
  value of \ paraphrased reference translations in parameter optimization.
\newblock In \emph{Proceedings of the Eighth Conference of the Association for
  Machine Translation in the Americas}, Waikiki, Hawaii.

\bibitem[{Marton et~al.(2009)Marton, Callison-Burch, and
  Resnik}]{Marton:2009:ISM:1699510.1699560}
Yuval Marton, Chris Callison-Burch, and Philip Resnik. 2009.
\newblock \href {http://dl.acm.org/citation.cfm?id=1699510.1699560} {Improved
  statistical machine translation using monolingually-derived paraphrases}.
\newblock In \emph{Proceedings of the 2009 Conference on Empirical Methods in
  Natural Language Processing: Volume 1 - Volume 1}, EMNLP '09, pages 381--390,
  Stroudsburg, PA, USA. Association for Computational Linguistics.

\bibitem[{Norouzi et~al.(2016)Norouzi, Bengio, Chen, Jaitly, Schuster, Wu, and
  Schuurmans}]{NIPS2016_6547}
Mohammad Norouzi, Samy Bengio, zhifeng Chen, Navdeep Jaitly, Mike Schuster,
  Yonghui Wu, and Dale Schuurmans. 2016.
\newblock \href
  {http://papers.nips.cc/paper/6547-reward-augmented-maximum-likelihood-for-neural-structured-prediction.pdf}
  {Reward augmented maximum likelihood for neural structured prediction}.
\newblock In D.~D. Lee, M.~Sugiyama, U.~V. Luxburg, I.~Guyon, and R.~Garnett,
  editors, \emph{Advances in Neural Information Processing Systems 29}, pages
  1723--1731. Curran Associates, Inc.

\bibitem[{Ott et~al.(2019)Ott, Edunov, Baevski, Fan, Gross, Ng, Grangier, and
  Auli}]{ott-etal-2019-fairseq}
Myle Ott, Sergey Edunov, Alexei Baevski, Angela Fan, Sam Gross, Nathan Ng,
  David Grangier, and Michael Auli. 2019.
\newblock \href {https://doi.org/10.18653/v1/N19-4009} {fairseq: A fast,
  extensible toolkit for sequence modeling}.
\newblock In \emph{Proceedings of the 2019 Conference of the North {A}merican
  Chapter of the Association for Computational Linguistics (Demonstrations)},
  pages 48--53, Minneapolis, Minnesota. Association for Computational
  Linguistics.

\bibitem[{Post(2018)}]{post-2018-call}
Matt Post. 2018.
\newblock \href {https://doi.org/10.18653/v1/W18-6319} {A call for clarity in
  reporting {BLEU} scores}.
\newblock In \emph{Proceedings of the Third Conference on Machine Translation:
  Research Papers}, pages 186--191, Brussels, Belgium. Association for
  Computational Linguistics.

\bibitem[{Rubino(2018)}]{rubino-2018-keynote}
Carl Rubino. 2018.
\newblock \href {https://www.aclweb.org/anthology/W18-1902} {{K}eynote: Setting
  up a machine translation program for {IARPA}}.
\newblock In \emph{Proceedings of the 13th Conference of the Association for
  Machine Translation in the {A}mericas (Volume 2: User Papers)}, Boston, MA.
  Association for Machine Translation in the Americas.

\bibitem[{Sennrich et~al.(2016)Sennrich, Haddow, and
  Birch}]{sennrich-etal-2016-improving}
Rico Sennrich, Barry Haddow, and Alexandra Birch. 2016.
\newblock \href {https://doi.org/10.18653/v1/P16-1009} {Improving neural
  machine translation models with monolingual data}.
\newblock In \emph{Proceedings of the 54th Annual Meeting of the Association
  for Computational Linguistics (Volume 1: Long Papers)}, pages 86--96, Berlin,
  Germany. Association for Computational Linguistics.

\bibitem[{Sennrich and Zhang(2019)}]{sennrich-zhang-2019-revisiting}
Rico Sennrich and Biao Zhang. 2019.
\newblock \href {https://doi.org/10.18653/v1/P19-1021} {Revisiting low-resource
  neural machine translation: A case study}.
\newblock In \emph{Proceedings of the 57th Annual Meeting of the Association
  for Computational Linguistics}, pages 211--221, Florence, Italy. Association
  for Computational Linguistics.

\bibitem[{Shen et~al.(2016)Shen, Cheng, He, He, Wu, Sun, and
  Liu}]{shen-etal-2016-minimum}
Shiqi Shen, Yong Cheng, Zhongjun He, Wei He, Hua Wu, Maosong Sun, and Yang Liu.
  2016.
\newblock \href {https://doi.org/10.18653/v1/P16-1159} {Minimum risk training
  for neural machine translation}.
\newblock In \emph{Proceedings of the 54th Annual Meeting of the Association
  for Computational Linguistics (Volume 1: Long Papers)}, pages 1683--1692,
  Berlin, Germany. Association for Computational Linguistics.

\bibitem[{Szegedy et~al.(2016)Szegedy, Vanhoucke, Ioffe, Shlens, and
  Wojna}]{labelsmooth}
Christian Szegedy, Vincent Vanhoucke, Sergey Ioffe, Jon Shlens, and Zbigniew
  Wojna. 2016.
\newblock \href {https://doi.org/10.1109/CVPR.2016.308} {Rethinking the
  inception architecture for computer vision}.
\newblock In \emph{2016 IEEE Conference on Computer Vision and Pattern
  Recognition (CVPR)}, pages 2818--2826.

\bibitem[{Thompson and Post(2020{\natexlab{a}})}]{thompson-post-2020-automatic}
Brian Thompson and Matt Post. 2020{\natexlab{a}}.
\newblock Automatic machine translation evaluation in many languages via
  zero-shot paraphrasing.
\newblock In \emph{Proceedings of the 2020 Conference on Empirical Methods in
  Natural Language Processing}, Online. Association for Computational
  Linguistics.

\bibitem[{Thompson and
  Post(2020{\natexlab{b}})}]{thompson-post-2020-paraphrase}
Brian Thompson and Matt Post. 2020{\natexlab{b}}.
\newblock Paraphrase generation as zero-shot multilingual translation:
  Disentangling semantic similarity from lexical and syntactic diversity.
\newblock In \emph{Proceedings of the Fifth Conference on Machine Translation
  (Volume 1: Research Papers)}, Online. Association for Computational
  Linguistics.

\bibitem[{Tiedemann(2012)}]{tiedemann-2012-parallel}
J{\"o}rg Tiedemann. 2012.
\newblock \href
  {http://www.lrec-conf.org/proceedings/lrec2012/pdf/463_Paper.pdf} {Parallel
  data, tools and interfaces in {OPUS}}.
\newblock In \emph{Proceedings of the Eighth International Conference on
  Language Resources and Evaluation ({LREC}'12)}, pages 2214--2218, Istanbul,
  Turkey. European Language Resources Association (ELRA).

\bibitem[{Vaswani et~al.(2017)Vaswani, Shazeer, Parmar, Uszkoreit, Jones,
  Gomez, Kaiser, and Polosukhin}]{transformer}
Ashish Vaswani, Noam Shazeer, Niki Parmar, Jakob Uszkoreit, Llion Jones,
  Aidan~N Gomez, {\L}ukasz Kaiser, and Illia Polosukhin. 2017.
\newblock \href
  {http://papers.nips.cc/paper/7181-attention-is-all-you-need.pdf} {Attention
  is all you need}.
\newblock In I.~Guyon, U.~V. Luxburg, S.~Bengio, H.~Wallach, R.~Fergus,
  S.~Vishwanathan, and R.~Garnett, editors, \emph{Advances in Neural
  Information Processing Systems 30}, pages 5998--6008. Curran Associates, Inc.

\bibitem[{Wang et~al.(2018)Wang, Pham, Dai, and
  Neubig}]{wang-etal-2018-switchout}
Xinyi Wang, Hieu Pham, Zihang Dai, and Graham Neubig. 2018.
\newblock \href {https://doi.org/10.18653/v1/D18-1100} {{S}witch{O}ut: an
  efficient data augmentation algorithm for neural machine translation}.
\newblock In \emph{Proceedings of the 2018 Conference on Empirical Methods in
  Natural Language Processing}, pages 856--861, Brussels, Belgium. Association
  for Computational Linguistics.

\bibitem[{Wieting et~al.(2019{\natexlab{a}})Wieting, Berg-Kirkpatrick, Gimpel,
  and Neubig}]{wieting-etal-2019-beyond}
John Wieting, Taylor Berg-Kirkpatrick, Kevin Gimpel, and Graham Neubig.
  2019{\natexlab{a}}.
\newblock \href {https://doi.org/10.18653/v1/P19-1427} {Beyond {BLEU}:training
  neural machine translation with semantic similarity}.
\newblock In \emph{Proceedings of the 57th Annual Meeting of the Association
  for Computational Linguistics}, pages 4344--4355, Florence, Italy.
  Association for Computational Linguistics.

\bibitem[{Wieting and Gimpel(2018)}]{wieting-gimpel-2018-paranmt}
John Wieting and Kevin Gimpel. 2018.
\newblock \href {https://doi.org/10.18653/v1/P18-1042} {{P}ara{NMT}-50{M}:
  Pushing the limits of paraphrastic sentence embeddings with millions of
  machine translations}.
\newblock In \emph{Proceedings of the 56th Annual Meeting of the Association
  for Computational Linguistics (Volume 1: Long Papers)}, pages 451--462,
  Melbourne, Australia. Association for Computational Linguistics.

\bibitem[{Wieting et~al.(2019{\natexlab{b}})Wieting, Gimpel, Neubig, and
  Berg-Kirkpatrick}]{wieting-etal-2019-simple}
John Wieting, Kevin Gimpel, Graham Neubig, and Taylor Berg-Kirkpatrick.
  2019{\natexlab{b}}.
\newblock \href {https://doi.org/10.18653/v1/P19-1453} {Simple and effective
  paraphrastic similarity from parallel translations}.
\newblock In \emph{Proceedings of the 57th Annual Meeting of the Association
  for Computational Linguistics}, pages 4602--4608, Florence, Italy.
  Association for Computational Linguistics.

\bibitem[{Wieting et~al.(2017)Wieting, Mallinson, and
  Gimpel}]{wieting-etal-2017-learning}
John Wieting, Jonathan Mallinson, and Kevin Gimpel. 2017.
\newblock \href {https://doi.org/10.18653/v1/D17-1026} {Learning paraphrastic
  sentence embeddings from back-translated bitext}.
\newblock In \emph{Proceedings of the 2017 Conference on Empirical Methods in
  Natural Language Processing}, pages 274--285, Copenhagen, Denmark.
  Association for Computational Linguistics.

\bibitem[{Williams and Zipser(1989)}]{doi:10.1162/neco.1989.1.2.270}
Ronald~J. Williams and David Zipser. 1989.
\newblock \href {https://doi.org/10.1162/neco.1989.1.2.270} {A learning
  algorithm for continually running fully recurrent neural networks}.
\newblock \emph{Neural Computation}, 1(2):270--280.

\bibitem[{Zhou et~al.(2019)Zhou, Ma, Hu, and Neubig}]{zhou-etal-2019-handling}
Chunting Zhou, Xuezhe Ma, Junjie Hu, and Graham Neubig. 2019.
\newblock \href {https://doi.org/10.18653/v1/D19-1143} {Handling syntactic
  divergence in low-resource machine translation}.
\newblock In \emph{Proceedings of the 2019 Conference on Empirical Methods in
  Natural Language Processing and the 9th International Joint Conference on
  Natural Language Processing (EMNLP-IJCNLP)}, pages 1388--1394, Hong Kong,
  China. Association for Computational Linguistics.

\end{thebibliography}
\bibliographystyle{acl_natbib}

\end{document}